\documentclass[english]{article}
\usepackage[]{fontenc}
\usepackage[latin1]{inputenc}
\usepackage{geometry}
\usepackage{amsmath}

\makeatletter

\providecommand{\LyX}{L\kern-.1667em\lower.25em\hbox{Y}\kern-.125emX\@}

\newcommand{\sig}{\makebox{$\hat{\sigma}^2$}}

\usepackage{babel}
\makeatother
\begin{document}

\title{Using Artificial Intelligence for Model Selection\date{}}

\author{Darin Goldstein \cr Department of Computer Science \cr California
State University, Long Beach \cr \textit{daring@cecs.csulb.edu}
\\
\\
Will Murray \cr  Department of Mathematics and Statistics \cr California
State University, Long Beach \cr \textit{wmurray@csulb.edu} \\
\\
Binh H. Yang \cr  Department of Epidemiology \cr University of California,
Los Angeles \cr School of Public Health \cr \textit{binhyang@ucla.edu}\\
}

\maketitle
\begin{abstract}
We apply the optimization algorithm Adaptive Simulated Annealing (ASA)
to the problem of analyzing data on a large population and selecting
the best model to predict the probability that an individual with
various traits will have a particular disease. We compare ASA with
traditional forward and backward regression on computer simulated
data. We find that the traditional methods of modeling are better
for smaller data sets whereas a numerically stable ASA seems to perform
better on larger and more complicated data sets.
\end{abstract}
Keywords: artificial intelligence, modeling, simulated annealing

\section{Introduction}

In this paper we apply a new method for model selection on large data
sets using the artificial intelligence algorithm Adaptive Simulated
Annealing (ASA). We focus here on an epidemiological setting, although
our techniques could be applied more widely. The objective is to analyze
a large amount of data on many different characteristics of a given
population and to select the model that best predicts the probability
that an individual with various traits will have a particular disease
outcome. The inclusion of categorical predictors that split up into
multiple {}``dummy'' variables and the possibility of cross terms
between the different characteristics make the number of variables
in a complete model prohibitively large. Thus it is necessary to select
a small number of variables that will give the most informative model.
Traditionally, this is done using forward regression or backward regression.
Instead, we place an upper bound on the number of variables we want
in our final model in advance and use ASA to decide which characteristics
(or combinations of characteristics) each variable should represent
to produce the model with the lowest $C_{p}$ statistic, a reflection
of bias and variance. We ran computer simulations on populations of
100,000, 500,000, and 1,000,000, equivalent to the data that might
be available for a major metropolitan area. Our results suggest that
this method produces a model with a $C_{p}$ statistic that is consistently
close to optimal whereas both forward and backward regression occasionally
do not.

\section{The Problem}

\label{section-problem}We assume that we have access to a large data
set with complete records of a population with a variety of fields,
including both continuous variables (such as age and alcoholic consumption)
and categorical variables (such as gender, ethnicity, and blood type).
We also assume that we know whether each individual in the population
has a particular disease, for example, liver cancer. (This is a binary
response, since the outcome is either {}``yes'' or {}``no'', but
our method could be easily modified to predict a continuous response,
for example, years of life lost due to liver cancer.) In our computer
simulations we used models of varying numbers of people with characteristics
randomly generated according to standard probabilities. We assigned
probabilities of getting the disease according to various risk factors
among the characteristics and let the computer randomly choose which
individuals to afflict. The characteristics we used were age, gender,
Hepatitis B viral infection, Hepatitis C viral infection, aflotoxin
exposure, genetic marker, alcohol, and tobacco. Simulations with 1,000,000
people, equivalent to a major metropolitan area, resulted in about
400-450 people getting the disease in any given trial. We then selected
an equal number of healthy people randomly from the population and
added these to the diseased people to form a data set for a simulated
population-based case-control study.

A classical problem of public health is then to develop a model from
the data set that predicts an individual's probability of getting
the disease based on his/her combination of characteristics. To do
this, we create a set of variables $\vec{w}=\{w_{i}\}$ as follows:
For each continuous characteristic (such as age or alcohol consumption)
we create one variable, and for each categorical characteristic (such
as gender or blood type) with $n$ categories, we create $n-1$ indicator
variables that can take values of 0 or 1. The $i$th indicator formed
from a particular categorical variable will indicate inclusion in
category $i$. (We only need $n-1$ of these indicator variables because
if all such variables are 0, then we are guaranteed inclusion into
the final category.) We then posit a model of the form 

\[
g(\vec{w})=\gamma _{0}+\sum _{i}\gamma _{i}w_{i}+\sum _{i<j}\gamma _{i,j}w_{i}w_{j}\]
in which the $\gamma _{i}$ and $\gamma _{i,j}$ are constant coefficients
and we omit the cross term $w_{i}w_{j}$ when $w_{i}$ and $w_{j}$
arise from the same categorical variable. This model is linear (in
the sense that if all variables but one are held constant, it is linear
in the nonconstant variable), but we could as easily include higher
degree terms and cross terms involving more than two variables if
necessary. In practice it is uncommon for biological models to include
terms involving three or more variables; we exclude these terms because
they are difficult to interpret biologically.

To simplify the notation, we can change variables from \textbf{$\vec{w}$}
to $\vec{x}$, where each $x_{i}$ represents either 1 (the constant
term), one of the $w_{i}$'s, or one of the $w_{i}w_{j}$'s. Then,
replacing the various $\gamma $'s with a single set of coefficients
$\{\beta _{i}\}$, the model above becomes \begin{equation}
g(\vec{w})=f(\vec{x})=\sum _{i}\beta _{i}x_{i}=\vec{\beta }\cdot \vec{x}\label{equation-model}\end{equation}
In the traditional approach to this problem, which, as we will see
below, uses forward or backward regression, the ordering of the $x_{i}$'s,
in other words, which $x_{i}$ is assigned to which $w_{i}$ or $w_{i}w_{j}$,
is not important. However, our method using Adaptive Simulated Annealing
depends heavily on finding a structured way of assigning the $x_{i}$'s
so that $x_{i}$ will have some relation to $x_{j}$ when $i$ is
close to $j$. We will discuss this further in Section \ref{section-algorithm}
below. 

One final modification is necessary to our model. To account for the
fact that the response from a known individual is either {}``healthy''
or {}``sick'', we can, at the end, transform the model (\ref{equation-model})
by the logistic function 

\[
P(\vec{x})=\frac{1}{1+e^{-f(\vec{x})}}\, .\]
For a detailed explanation of this transformation, see Section 12.12
in \cite{walpole}. 

The problem now is that of economy of terms; in our simulation, for
example, the eight characteristics we studied produced 61 variables,
$x_{0}$ to $x_{60}$. We therefore try to find a model using fewer
terms and identify the $x_{i}$'s (and hence the individual characteristics
or combinations thereof) that produce the most informative results.
A commonly used measure of the efficiency of a model is the $C_{p}$
statistic 

\[
C_{p}=p+\frac{(s^{2}-\sig )(n-p)}{\sig },\]
where $p$ is the number of variables in the current model, $s^{2}$
is the mean square error, and $n$ is the total possible number of
variables. Finally, $\hat{\sigma }^{2}$ is an estimate of $\sigma ^{2}$,
the error variance in the population; since this latter quantity is
unavailable, we take $\hat{\sigma }^{2}$ to be the mean square error
for the most complete model, i.e. the one incorporating all the variables.
Then a lower $C_{p}$ value indicates a more desirable model. For
a general overview of these terms, see Chapters 11 and 12 in \cite{walpole};
for a more detailed development of the the $C_{p}$ statistic, see
\cite{myers}. 

Two traditional methods of solving this problem are forward regression
and backward regression. In the former, one starts with only the constant
variable and adds variables to the model one at a time, each time
selecting the variable that lowers the $C_{p}$ statistic the most.
The process terminates when no variables can be added that lower the
statistic. Backward regression is similar, starting with all the variables
and progressively deleting them. Both procedures are computationally
intensive, since they involve at each step computing a least squares
solution to $\textbf X\vec{\beta }=\vec{y}$ (where $\textbf X$ represents
the data matrix) for each variable that is under consideration to
be added or deleted from the model. This in turn requires finding
the inverse (or pseudoinverse) of the matrix ${\textbf X}^{T}{\textbf X}$,
which has dimension equal to the number of variables currently under
consideration (up to 61x61 in our simulations). As greedy algorithms,
they are also not guaranteed to produce an optimal solution. 

As we noted above, since forward and backward regression test at each
step every remaining variable to see which one should be added or
deleted, these methods are not dependent on the ordering of the variables
$x_{i}$. 

Our new method described in Section~\ref{section-algorithm} below
is to determine an appropriate number of variables in advance and
then to use ASA to determine which variables to include in the model
to yield the lowest $C_{p}$ statistic. We will see that the ordering
of the $x_{i}$'s is important to this method.

\section{Adaptive Simulated Annealing}

Simulated annealing was introduced in 1983 by Kirkpatrick, Gelatt,
and Vecchi (\cite{kirkpatrick}) as a method of function optimization
that is particularly well suited to functions that are difficult to
evaluate in any continuous manner. It is based on the physical process
of annealing, in which the molecules of a material are brought into
a crystalline arrangement by gradually cooling the material. Since
the crystal is the most ordered configuration of the molecules, it
is the one that minimizes the total energy of the system, so the molecules
should ultimately come to rest in that configuration. However, there
may be other arrangements of the molecules that are stable despite
having a higher total energy than the minimum (i.e. local minima of
the energy function), and to avoid the system coming to rest in one
of these arrangements the scientist must be careful to give the system
enough initial energy and to avoid cooling it too quickly. At each
stage of the cooling, small temperature fluctuations within the material
will create and destroy defects until equilibrium for that temperature
is achieved. 

Simulated annealing is a procedure for minimization of a function
of several variables in which the values of the variables represent
configurations of the molecules of the system and the objective function
represents its total energy. The computer initially assigns random
values to the variables and then gives the system a certain {}``temperature'',
i.e. a tendency of the variables to move randomly. Each move will
affect the total energy of the system as measured by the objective
function, and the temperature is used to determine the probability
that a move that raises the total energy will be accepted. After enough
moves have been made to simulate the equilibrium activity of the material
at that temperature, the temperature of the system is lowered and
the process begins again. Eventually, the temperature is low enough
that the variables no longer change significantly and a minimum is
achieved. It is important to note, however, that although physical
annealing is known in theory to produce the global minimum of the
total energy of a system, the efficiency of simulated annealing depends
on the function being minimized and it is difficult to guarantee that
a given annealing schedule will produce a global minimum rather than
a local one. 

In our investigation we use an adaptive version of simulated annealing
introduced by Lester Ingber (\cite{ingbercode}, \cite{ingberasa},
\cite{ingberasa2}, \cite{ingberasa3}) as an improvement on his earlier
algorithm for Very Fast Simulated Reannealing (\cite{ingbervfsr},
\cite{ingbervfsr2}). ASA takes advantage of structure on the input
of the objective function (in our case, the ordering of the $x_{i}$'s
described below in Section~\ref{section-algorithm}) to decrease
the running time of the algorithm and increase the probability that
it will find a global minimum of the function. Optimization using
VFSR and ASA has been effectively applied in wide variety of situations,
including three dimensional image compression (\cite{forman}, \cite{forman2}),
modeling of financial markets (\cite{ingberfinancial}, \cite{ingberfinancial2},
\cite{ingberfinancial3}, \cite{ingberfinancial4}, \cite{ingberfinancial5},
\cite{sakata}, \cite{sakata2}), dairy farming (\cite{mayer}, \cite{mayer2}),
neural networks (\cite{cozzio}, \cite{cohen}, \cite{indiveri},
\cite{ingberneural}), geophysical inversion (\cite{sen}), magnetic
resonance imaging (\cite{buszko}), electroencephalography (\cite{ingberelectro},
\cite{ingberelectro2}), and combat simulation (\cite{bowman}, \cite{ingbercombat}).
An article on the many applications of ASA has appeared in The Wall
Street Journal (\cite{wofsey}).

\section{The Algorithm}

\label{section-algorithm}

The point at which our new method departs from traditional solutions
to this problem is in the variable change from the $w_{i}$'s to the
$x_{i}$'s in the model (\ref{equation-model}) above. As we have
seen at the end of Section~\ref{section-problem}, the efficiency
of forward and backward regression is not dependent on the ordering
of the $x_{i}$'s. In our method, however, we use the following {}``zipper''
algorithm to ensure that if $i$ is close to $j$, then $x_{i}$ and
$x_{j}$ represent similar values of the $w$'s: Suppose, for example,
that there are only four $w_{i}$'s. Then the $x_{i}$'s would be
assigned as follows: 

\[
\begin{array}{llll}
 x_{0}:=1, &  &  & \\
 x_{1}:=w_{1}, & x_{2}:=w_{1}w_{2}, & x_{3}:=w_{1}w_{3}, & x_{4}:=w_{1}w_{4},\\
 x_{5}:=w_{4}, & x_{6}:=w_{3}w_{4}, & x_{7}:=w_{2}w_{4}, & \\
 x_{8}:=w_{2}, & x_{9}:=w_{2}w_{3}, &  & \\
 x_{10}:=w_{3}. &  &  & \end{array}
\]
(Here we have assumed that there were no categorical characteristics
with more than two categories, so that no $w_{i}$ and $w_{j}$ arose
from the same characteristic. If they had, then the corresponding
cross term $w_{i}w_{j}$ would be omitted from the assignment above.)
By using this algorithm, we attempt to structure the search space
in such a way as to make the search for the optimal solution as easy
as possible for ASA. To get an idea of why this structuring of the
search space is important, consider the difference between trying
to find the minimum of a straight line and trying to find the minimum
of a series of randomly placed points in the plane. Obviously, the
former is much easier (both for a human and a computer). However,
the former situation can become the latter very quickly if we allow
the points along the $x$-axis to shuffle themselves randomly (thereby
causing the $x$-coordinates of the line to shuffle themselves randomly
and removing the ordered structuring of the search space). In the
model above, by guaranteeing that $x_{i}$ and $x_{i+1}$ always contain
exactly one of the $w_{j}$'s in common, we hope that the $C_{p}$
value does not change too radically when transitioning between the
two. Of course, we can make no guarantees.

We decide in advance the number $p$ of the $x_{i}$'s we want our
model to include. This could be any number from 0 to $n$, but we
found in our simulations with $n=60$ that $p\approx 5-10$ was usually
optimal. This range is based on the outcomes of forward and backward
regression, on trial and error with our ASA program, on running time
(trials with ten variables took about ten minutes on a Pentium 4 850
MHz processor), and on current practice among researchers in epidemiology. 

We then set up dummy variables $z_{i},1\leq i\leq p$, which may take
integer values between 1 and $n$ depending on which of the $x_{i}$'s
would be included in the model. Our goal is then to find the values
of the $z_{i}$'s that minimize the objective function defined by
the $C_{p}$ statistic for a given model. This is where we use simulated
annealing. 

The publicly available C-code for ASA (\cite{ingbercode}) allows
the user to define how to evaluate the objective function to be minimized.
In our case, evaluation on a particular set of values of the $z_{i}$'s
required building a model with the corresponding $x_{i}$'s included,
finding the coefficients $\beta _{i}$ by calculating a least squares
solution to $\textbf X\vec{\beta }=\vec{y}$, and calculating the
$C_{p}$ statistic for that model. We modified the public code accordingly
and ran multiple simulations with various sizes of data sets.

\section{The Results}

We ran several simulations on computer-generated data. In all cases,
we used our ASA method and also ran forward and backward regression
on the data as a control. We compared the three methods according
to the final $C_{p}$ statistic produced and the running time required
to achieve it as measured by the number of function evaluations involved.
Here one function evaluation is defined to be the calculation of the
$C_{p}$ statistic for a given set of variables as described at the
end of Section \ref{section-algorithm}, since forward and backward
regression use the same procedure as they search for variables to
be added to or deleted from the model. It is true that not all function
evaluations will take equal amounts of processor time, since an evaluation
with more variables will involve finding the pseudoinverse of a larger
matrix; however, we believe, in keeping with standard practice in
computer science, that such a measure is still meaningful enough to
merit study. 

We show below the results of our simulations on computer-generated
data. Using the characteristics described above in Section \ref{section-problem},
there were 60 possible variables; we set our ASA program to search
for the best seven variables. We ran five simulations each on populations
of size 100,000, 500,000, and 1,000,000 with the following results.
\vfill \eject

\begin{center}Simulated populations of 100,000 (approximately 45 diseased,
45 healthy studied). \\
\end{center}

\begin{center}\begin{tabular}{|l|l|l|l|}
\hline 
\multicolumn{1}{|c||}{}&
\multicolumn{1}{c||}{Forward Regression}&
\multicolumn{1}{c||}{Backward Regression}&
\multicolumn{1}{c|}{ASA}\\
\hline
\hline 
 $\begin{array}{l}
 C_{p}\mbox {\, statistic}\end{array}
$&
 $\begin{array}{l}
 -46.387\\
 -45.831\\
 -54.046\\
 -56.199\\
 -54.936\end{array}
$&
 $\begin{array}{l}
 -29.574\\
 -43.318\\
 -50.251\\
 -56.199\\
 -49.022\end{array}
$&
 $\begin{array}{l}
 -46.734\\
 -48.968\\
 -53.009\\
 -56.199\\
 -54.724\end{array}
$\\
\hline
\hline 
 $\begin{array}{l}
 \mbox {Function\, evaluations}\end{array}
$&
 $\begin{array}{l}
 177\\
 119\\
 177\\
 119\\
 177\end{array}
$&
 $\begin{array}{l}
 1725\\
 1802\\
 1824\\
 1830\\
 1820\end{array}
$&
 $\begin{array}{l}
 1341\\
 1677\\
 1587\\
 2827\\
 199\end{array}
$\\
\hline
\end{tabular}\end{center}

\begin{center}Simulated populations of 500,000 (approximately 225
diseased, 225 healthy studied).\\
\end{center}

\begin{center}\begin{tabular}{|l|l|l|l|}
\hline 
\multicolumn{1}{|c||}{}&
\multicolumn{1}{c||}{Forward Regression}&
\multicolumn{1}{c||}{Backward Regression}&
\multicolumn{1}{c|}{ASA}\\
\hline
\hline 
 $\begin{array}{l}
 C_{p}\mbox {\, statistic}\end{array}
$&
 $\begin{array}{l}
 -47.511\\
 -43.141\\
 -50.803\\
 -36.406\\
 -45.648\end{array}
$&
 $\begin{array}{l}
 -41.831\\
 -37.691\\
 -41.506\\
 -35.767\\
 -41.693\end{array}
$&
 $\begin{array}{l}
 -47.510\\
 -43.152\\
 -51.160\\
 -41.127\\
 -46.063\end{array}
$\\
\hline
\hline 
 $\begin{array}{l}
 \mbox {Function\, evaluations}\end{array}
$&
 $\begin{array}{l}
 119\\
 399\\
 290\\
 119\\
 452\end{array}
$&
 $\begin{array}{l}
 1815\\
 1785\\
 1764\\
 1739\\
 1752\end{array}
$&
 $\begin{array}{l}
 2442\\
 3336\\
 1305\\
 3780\\
 2994\end{array}
$\\
\hline
\end{tabular}\end{center}

\begin{center}Simulated populations of 1,000,000 (approximately 450
diseased, 450 healthy studied).\\
\end{center}

\begin{center}\begin{tabular}{|l|l|l|l|}
\hline 
\multicolumn{1}{|c||}{}&
\multicolumn{1}{c||}{Forward Regression}&
\multicolumn{1}{c||}{Backward Regression}&
\multicolumn{1}{c|}{ASA}\\
\hline
\hline 
 $\begin{array}{l}
 C_{p}\mbox {\, statistic}\end{array}
$&
 $\begin{array}{l}
 -47.162\\
 -44.149\\
 -40.755\\
 -16.518\\
 -43.012\end{array}
$&
 $\begin{array}{l}
 -39.287\\
 -41.134\\
 -21.738\\
 -22.845\\
 -40.003\end{array}
$&
 $\begin{array}{l}
 -48.347\\
 -44.148\\
 -40.743\\
 -46.679\\
 -38.235\end{array}
$\\
\hline
\hline 
 $\begin{array}{l}
 \mbox {Function\, evaluations}\end{array}
$&
 $\begin{array}{l}
 234\\
 345\\
 234\\
 177\\
 504\end{array}
$&
 $\begin{array}{l}
 1764\\
 1764\\
 1659\\
 1620\\
 1752\end{array}
$&
 $\begin{array}{l}
 461\\
 329\\
 678\\
 690\\
 1913\end{array}
$\\
\hline
\end{tabular}\vfill \eject\end{center}

Asymptotically, the number of function evaluations necessary to compute
forward regression is easily seen to be $O(nk)$ where $n$ represent
the total number of variables under consideration and $k$ represents
the final number of variables chosen in the model. It is a notoriously
difficult problem of theoretical computer science to analyze the computational
complexity of ASA \cite{jerrum-sorkin,mrsv,sorkin}. We can only present
our results and leave it to the interested reader to draw his own
conclusions.

\section{Advantages and Limitations}

\label{section-advantages}We found that running ASA in most cases
produces a slightly lower $C_{p}$ statistic especially on the trials
with large data sets. The cases in which ASA did not produce a lower
$C_{p}$ statistic were ones in which the optimal solution required
significantly more or fewer than the seven variables we programmed
ASA to search for; in these cases the $C_{p}$ statistic produced
by ASA was still close to that of the other methods, and when we reconfigured
ASA to search for an appropriate number of variables it produced a
lower $C_{p}$ statistic. It appears that for small data sets with
few parameters, the traditional methods of modeling are slightly preferable
to ASA. The benefits of ASA seem to be more pronounced for larger
data sets.

Another advantage of ASA is that its results seem to be more predictable
than those of the other methods. Although in many tests the results
of the three methods were comparable, in several instances forward
regression and backward regression produced $C_{p}$ statistics that
were considerably higher than the lowest obtained $C_{p}$ statistic;
the statistics produced by ASA were never far from the best statistics
produced by the three methods. 

One disadvantage of our method is that it requires the user to determine
the number of variables in the model beforehand. Simulations run under
the same conditions can have optimal solutions (as determined by forward
and backward regression) requiring quite different numbers of variables;
in our simulations with 60 possible nonconstant variables we found
some for which the $C_{p}$ statistic was optimized with just one
variable and some requiring as many as 21. 

In practice this should not be a serious concern for three reasons.
One is that a researcher in epidemiology will usually have a rough
idea of the desired number of variables in advance as a compromise
between the needs of the problem and the availability of computing
power. (The assumption above that the model is linear already requires
that the researcher know in advance something about the desired model.)
A second reason is that as part of the random component of the program
it automatically checks the possibility of deleting variables from
the model, so if the minimum involves fewer variables than the user
selects, the program will still find it. Finally, this property may
even be considered an advantage, since our method allows the user
to control the sophistication of the model (i.e. the number of variables
it will use) in advance, whereas with forward and backward regression
the models produced may have widely varying numbers of variables. 

Another disadvantage of our method is that it is not guaranteed to
find the global minimum of the function we are trying to optimize.
Neither, however, do the traditional methods of regression, and indeed,
such a goal would be impractical because of the number of parameters
in the problem.

\section{Further Research}

Though the results of the computer simulations are suggestive, we
would like to see more done in the area of practical epidemiology,
perhaps a large-scale analysis of publicly available data sets. Such
an undertaking is non-trivial, however, even given this previous research.
Most publicly available data sets are enormous in comparison with
the relatively small ones used in this study (several hundreds versus
over fifteen thousand). Also, the number of possible choices for variables
is immense. The investment of computer time would be substantial even
for a modestly ambitious study. An even greater hurdle seems to be
the numerical stability of the algorithm itself. For data sets that
are extremely large or that have a very wide range of data values,
finding the least squares solution (which involves inverting a matrix)
is inherently a numerically unstable undertaking. We have attempted
programming the bulk of the algorithm in both C and Mathematica. In
C, the numerical stability of the unusually large number of computations
seems to be an issue; whereas, in Mathematica, even though the kernel
controls the numerical precision well, the time necessary for even
a relatively small computation is enormous.

\bibliographystyle{plain}
\bibliography{newepi}

\end{document}